\documentclass[runningheads]{llncs}
\usepackage{graphicx}
\usepackage{amsmath}
\usepackage{amsfonts}
\newcommand\myparagraph[1]{\vspace{12pt}\parindent0pt\textbf{#1}\hspace{8pt}}
\newcommand\etal{\textit{et al.}\enspace}
\newcommand\cf{\textit{cf.}\enspace}
\newcommand\eg{\textit{e.g.}\enspace}
\newcommand\ie{\textit{i.e.}\enspace}
\newcommand\sectionname{Sect.}

\begin{document}
\title{Evaluating Zero-cost Active Learning for Object Detection\thanks{Supported by Investitionsbank Berlin, Germany and computational resources of the BMBF grant programme ``KI-Nachwuchs@FH''.}}
\titlerunning{Evaluating Zero-cost Active Learning for Object Detection}
%
%
\author{Dominik Probst\inst{1} \and
Hasnain Raza\inst{2} \and
Erik Rodner\inst{1}\orcidID{0000-0002-3711-1498}}
\authorrunning{Dominik Probst, Hasnain Raza, and Erik Rodner}
%
\institute{KI-Werkstatt / Ingenieurinformatik, Fachbereich 2,\\ University of Applied Sciences, Berlin, Germany,\\ \email{erik.rodner@htw-berlin.de} \and
Hasty GmbH, Berlin, Germany}
\maketitle              
\begin{abstract}
Object detection requires substantial labeling effort for learning robust models. Active learning can reduce this effort by intelligently selecting relevant examples to be annotated. However, selecting these examples properly without introducing a sampling bias with a negative impact on the generalization performance is not straightforward and most active learning techniques can not hold their promises on real-world benchmarks.
In our evaluation paper, we
focus on active learning techniques without a computational overhead besides inference, something we refer to as zero-cost active learning. In particular, we show that a key ingredient is not only the score on a bounding box level but also the technique used for aggregating the scores for ranking images. We outline our experimental setup and also discuss practical considerations when using active learning for object detection.
\keywords{Active learning \and Object Detection \and Evaluation Paper}
\end{abstract}
\section{Introduction}
When creating machine learning models, one is often faced with the problem that although enough data is available, a large part of it is not annotated. To reduce the amount of annotated data required, various types of methods can be used: semi-supervised learning~\cite{zheng2022simmatch} for integrating unlabeled examples, weakly-supervised learning~\cite{reiss2021every} for using cheap annotation types, transfer learning~\cite{rodner2010one}, or domain adaptation~\cite{rodner2013towards} to exploit information from related tasks and data distributions. However, we focus on an orthogonal technique called active learning (AL), where an algorithm suggests informative samples to be labeled by an oracle (\textit{e.g.}, a human) which will likely yield the highest gain in model quality once being annotated and used for training. 

In contrast to a majority of AL literature that focuses on classification tasks~\cite{lewis1994heterogeneous,lewis1994sequential,joshi2009multi,Freytag14_SIE}, our application scenario is object detection, where annotation is even more costly. In addition, object detection is highly relevant for industrial visual inspection or autonomous mobility. In our work, we concentrate on AL methods that do not require changing the model
architecture and where scores can be calculated with a low computational overhead, something we define as \emph{zero-cost active learning} in this paper. With these practical constraints, a large number of active learning approaches are simply impossible to apply, since they require either an architectural change or a large additional computational overhead~\cite{yu2022consistency}.

\section{Related Work} 
While a lot of the literature focuses on active learning for image classification, it is difficult to find previous research that explore how these simple methods for uncertainty sampling perform on more complicated tasks like object detection, let alone their aggregation strategy. Furthermore, due to the rather low annotation cost of most classification tasks, active learning often does not yield performance benefits in practise. The publications of \cite{lewis1994sequential} and \cite{lewis1994heterogeneous} introduce uncertainty sampling. Joshi~\etal~\cite{joshi2009multi} explore uncertainty sampling for active learning by using the entropy as an uncertainty measure. Further works have extended these ideas to utilize other uncertainty measures such as variation ratios \cite{freeman1965elementary}, mean standard deviation \cite{BMVC2017_57}, margin \cite{roth2006margin}, and variance of the class probability distributions. The paper of Sener and Savarese~\cite{sener2017active} is interesting as it specifically tackles the active learning problem for convolutional neural networks~\cite{lecun1995convolutional} (relevant for object detection) and formulates it as a core set selection problem. 

A lot of these ideas have similarly been combined with Bayesian neural networks (\cf \cite{gal2016uncertainty} for a primer). Since Bayesian statistics are often approximated with multiple forward passes, using them in AL is often too time-consuming. For object detection, using the scores of all object instances in the image can be used as a surrogate statistic. This is precisely the setting we explore. Choi~\etal~\cite{choi2021active} presents active learning specifically for object detection, but their method involves modifying the detector network with mixture density networks for the localization and classification heads. Agarwal~\etal~\cite{agarwal2020contextual} explores a distance measurement based ``contextual diversity". The work of Yuan~\etal~\cite{yuan2021multiple} is similar to our approach since images are considered as ``bags of instances". However, their method makes modifications to the detector model, impractical for several applications. The same holds for \cite{roy2018deep}, although their setup is similar to ours except for the white boxing approach. Brust~\etal~\cite{brust2018active} is comparable to the methodology and idea of our work, the only difference being that we benchmark more active learning scoring functions.
The paper of \cite{feng2022albench} presents a framework for benchmarking active learning algorithms. Our underlying software design is 
similar, but it's presentation is not in the scope of this paper.

\section{Zero-cost Active Learning for Object Detection}
\label{sec:techniques}
Our setup for active learning is as follows: first, we assume an initial annotated training set of minimal size or directly a pre-trained detector model (already on the target task) to be given. Furthermore, we assume to have access to a large unlabeled dataset $\mathcal{D}_u = \left\{ I_k \right\}_k$. 
The detector is applied to all images of this dataset. However, if this is computational infeasible a random sampling could be applied to reduce the set, although we do not investigate this possibility here.

After applying the detector, we get multiple bounding boxes $B_{k,i}$ with  $1 \leq i \leq M_k$ for every image $I_k$, with each having class probabilities $p_{k,i,d}$, for all classes $1\leq d \leq D$. 
Selecting the most relevant examples to be labeled is then done by scoring each image $I_k$ with a function $s$ and finally providing the $L$ images with the highest score to the oracle. In a real-world scenario, the oracle would be a human annotator, while in our experimental setup 
we directly use the ground-truth of the benchmark dataset to automate the evaluation process.

In the following, we outline the different options for the scoring function $s$, which is calculated by first scoring bounding boxes and accumulating these individual scores for each image.

\subsection{Scoring bounding boxes}

The majority of the research done for active learning focuses on image classification. Therefore, there are multiple established methods directly
using inferred class probabilities $p$ and are thus suitable to define a scoring function $x(k,i)$ for a bounding box $i$ in image $k$.  

In the following, we skip the index $k$ of the image for notational brevity.

\myparagraph{Margin Score for Active Learning} The margin method of \cite{roth2006margin}, in other literature sometimes referred to as $1$-vs-$2$ method, is a measure for the uncertainty of the predicted probabilities. For a bounding box $B_i$ it is defined by:
\begin{align}
x_{\text{ms}} (i) &= 1-(p_{i,d_1}-p_{i,d_2} )
\end{align}
with $d_1$ being the class with the highest and $d_2$ being the class with the second highest predicted probability, respectively. 
If the detector is uncertain, at least two classes will be likely predicted with a similar probability. This results in a high value of $x_{\text{ms}}$ close to $1$. In rather certain cases, we will observe lower values of the score.

\myparagraph{Variance Score for Active Learning}
An alternative for the margin score, is to measure the variance of the predicted probabilities directly:
\begin{align}
x_{\text{vs}}(i) &= 1 - \frac{1}{D-1} \sum\limits_{d=1}^D \left(p_{i,d}- \frac{1}{D} \sum\limits_{j=1}^D p_{i,j} \right)^2 \enspace.
\end{align}
The reasoning for this measure is the same as the one for the margin score with the difference being that it does not only focus on the two most likely classes being predicted. 

\myparagraph{Entropy Score for Active Learning}
Since the output of the detector is a probability vector, we should more consequently 
use entropy as a measure of uncertainty:
\begin{align}
x_{\text{ent}}(i) &= - \left( \sum\limits_{d=1}^D p_{i,d} \cdot \log_2 p_{i,d} \right) / \log_2(D)\enspace.
\end{align}
Similarily to both previous scores, this score has its maximum at 1 and we are interested in the examples with the highest score.

\myparagraph{Random} To compare with an iterative annotation cycle that does not use active learning at all (passive learning), all images for the next learning process can also be selected with a random scoring function. This baseline has to be evaluated in any active learning benchmark, because none of the state-of-the-art active learning methods can guarantee any benefit at all compared to classical passive learning without an intelligent selection of examples during annotation.

\subsection{Scoring images by accumulating bounding box scores}
\label{sec:accumulation}
Since several objects can be detected for one image, the values $x$ of the individual detections must be accumulated to result in a score $s$ for the whole image instead of a single bounding box. Please note that in the following, $x(k,i)$ denotes an arbitrary active learning score (margin, variance, entropy, or random) of a bounding box $B(k,i)$. In our work, we use one of the following accumulation functions:

\myparagraph{Mean accumulation}
A straightforward way to accumulate bounding box active learning scores is to calculate the mean of all the scores for an image $I_k$:
\begin{align}
    s_{\text{mean}} (k) &= \frac{1}{M_k} \sum\limits_{i=1}^{M_k} x(k,i)\enspace.
\end{align}
Therefore, all bounding boxes affect the resulting score, also the ones that were detected with high confidence. This might be disadvantageous in the following scenario: let's assume that one of the existing object categories of the task is rather easy to detect with multiple instances in nearly every image, \eg, a face category. A detector that has very well learned to detect faces but confuses and misses all other categories, would still lead to a low active learning score for nearly all unlabeled images, despite the fact that they contain relevant instances of other object categories that should be annotated. 

\myparagraph{Sum accumulation}
An alternative is to compute the sum of all bounding box active learning scores:
\begin{align}
    s_{\text{sum}} (k)= \sum\limits_{i=1}^{M_k} x(k,i)\enspace.
\end{align}
The result does strongly depend on the number of objects in an image. This can have the effect that images with many objects achieve a higher score than images with fewer objects. For active learning, this can be reasonable, since the annotator automatically focuses on images with many objects in scenarios that are likely more challenging (due to overlaps, different sizes, etc.). However, it would also prevent the annotator from seeing images with object categories that usually appear in isolated environments.

\myparagraph{Maximum accumulation}
Given the obvious caveats of the previous methods, we can also compute the maximum bounding box active learning score of the whole image:
\begin{align}
s_{\text{max}} (k)=\max\limits_{1 \leq i \leq M_k} x(k,i) \enspace.
\end{align}
Consequently, the active learning score of an image only depends on a single most uncertain object detection. 

\section{Experiments and Evaluation}
In the following, we evaluate all resulting active learning approaches, \ie all combinations of bounding box scores $x$ and accumulation functions $s$.

\subsection{Design of the experiments}

First, we elaborate on our experimental setup including data, detector, and evaluation criterion used.

\myparagraph{Dataset}
Running active learning experiments requires hundreds of training runs to be performed. We therefore choose the Pascal VOC 2012~\cite{everingham2015pascal} dataset with the usual splits for experiments, since it has a medium size allowing several experiments on a standard GPU workstation (with two NVIDIA RTX 3090) and is non-trivial for object detection. The dataset contains several thousand images with overall twenty object categories. We use the validation dataset of Pascal VOC 2012 for evaluating the models. The Pascal VOC 2012 training set is used for the initial random training set as well as for the images the active learning techniques can select from.  

\myparagraph{Detector}
As a detector, we use a Faster R-CNN model~\cite{ren2015faster} (ResNet-50, FPN) since it is prevalent in many standard object detection implementations in industry. In particular, we used the implementation of detectron2~\cite{wu2019detectron2} and the standard training scheme implemented therein. All of our training runs had a length of $100$ epochs with early stopping.

\myparagraph{Evaluation criteria}
To compensate for deviations, the active learning process (all cycles) was repeated five times and the performance was determined using the mean average precision (mAP) of all categories with an intersection over union (IoU) threshold of 0.5 following the standard evaluation procedure for Pascal VOC 2012. Our oracle always annotated the given images with all ground-truth bounding boxes (see below for an discussion on oracle assumptions).

\myparagraph{Experimental setup}
The model was initially trained with 10 fully annotated images. Afterwards, an active learning technique was used to score each image. In active learning cycle $\ell$, the $L = 10\ell + 10$ images with the largest score were then given to the oracle to be labeled and used in the next cycle for training. Note that the number of training examples annotated increases over time.

The training data set was initially started with randomly selected labelled images and then steadily increased. 

\myparagraph{Oracle design}
It is important to note that we use a specific design of the oracle in our experiments. The oracle simulates a human annotator and should be adapted to the application under consideration. In our case, we make the following assumptions:
\begin{enumerate}
\item Given an image, the annotator annotates all ground-truth object instances present in the given images.
\item The annotator annotates perfect bounding boxes.
\item An annotator is provided with several images at once and all images
are annotated before the next active learning cycle $\ell$ continues.
\end{enumerate}

Assumption 1 and 2 definitely do not hold in practice. However, breaking these assumptions results in label noise for object detection as already studied in several works, such as \cite{adhikari2021effect}. We therefore assume perfect annotations to reduce the complexity of the experiments and since we did not see any surprising insights under noise influence in preliminary experiments. Selecting and annotating multiple images at once (assumption 3) is a batch active learning setting~\cite{citovsky2021batch}. There are multiple approaches tackling this scenario~\cite{zhdanov2019diverse} for classification tasks, since the relevance of the examples in a batch is not independent from each other.
All images in a batch might be individually relevant to be labeled but might be highly redundant when all of them are similar in the batch. In our evaluation, batch active learning is not considered, since it would involve an additional computational cost currently not feasible in practice.

\subsection{Evaluation}

\myparagraph{Quantitative evaluation}
\begin{figure}[tbp]
    \centering
    \includegraphics[width=\linewidth]{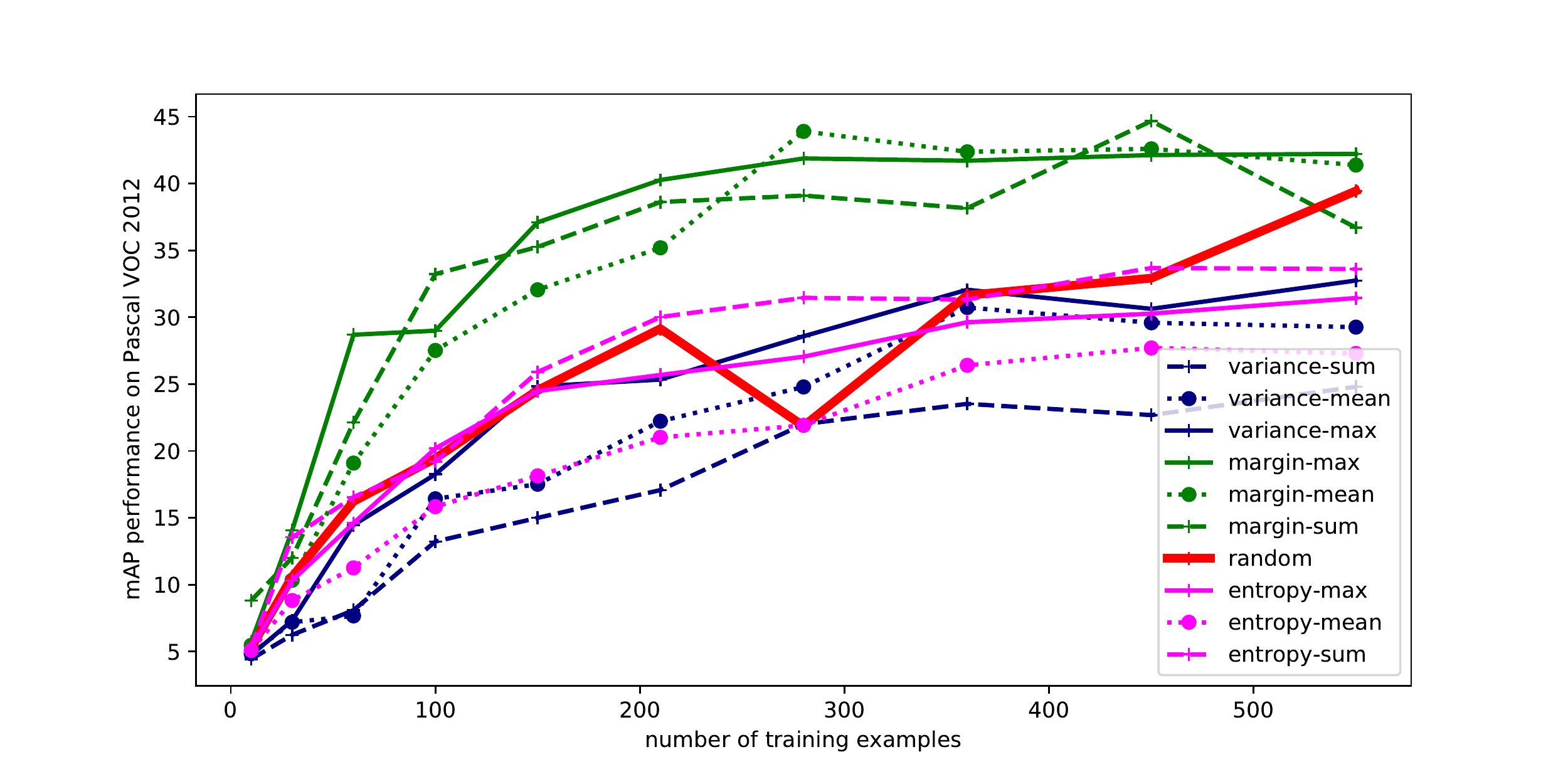}
    \caption{Performance of the different AL models with an increasing number of training examples.}
    \label{fig:increasingex}
\end{figure}
\begin{figure}[tbp]
    \centering
    \includegraphics[width=\linewidth]{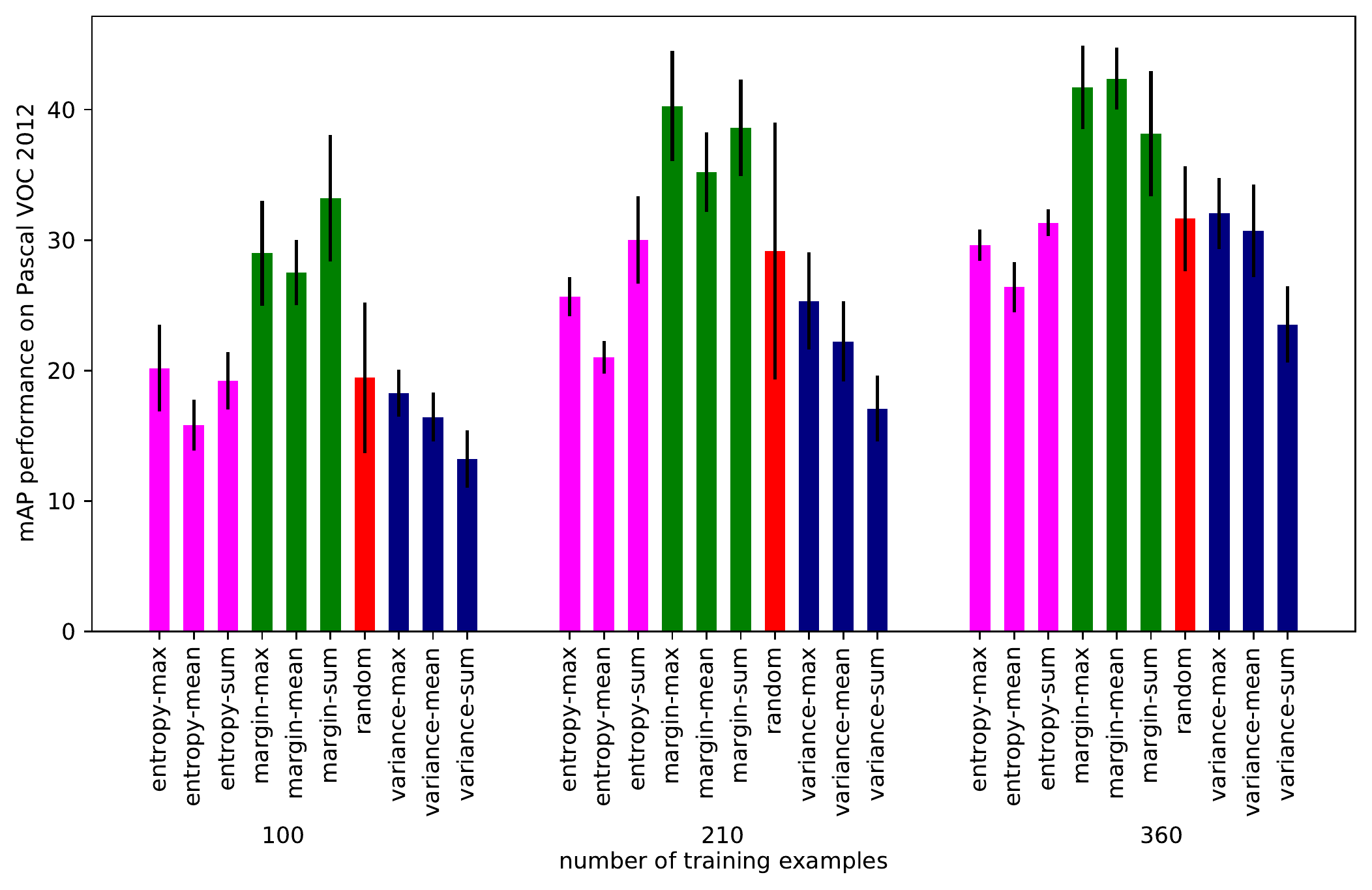}
    \caption{Performance of the different AL models for different numbers of training examples. The  standard deviation resulted from different data splits.}
    \label{fig:barplot}
\end{figure}
Our main results are given in \figurename~\ref{fig:increasingex}, where we plot mAP performance of the resulting detector with respect
to the number of training examples used. This plot shows the results for all method combinations (selection of a bounding box score method and
accumulation technique).

It can be seen that the margin criterion performs best in combination with a maximum accumulation. The choice of accumulation is indeed relevant, since mean accumulation of the margin scores results in significantly worse models with respect to mAP performance. Furthermore, the margin criterion is, irrespective of the accumulation method, the scoring technique of choice, since all studied alternatives (entropy and variance) are often even worse than
classical passive learning (denoted as random in the plot). The results shown in \figurename~\ref{fig:increasingex} hide the variation of the
model performance, \ie the standard deviation of the mAP values. Therefore, \figurename~\ref{fig:barplot} shows three points
in time (related to the number of training examples used) in detail with error bars.

\myparagraph{Qualitative evaluation}
\begin{figure}[tbp]
    \centering
    \includegraphics[width=\linewidth]{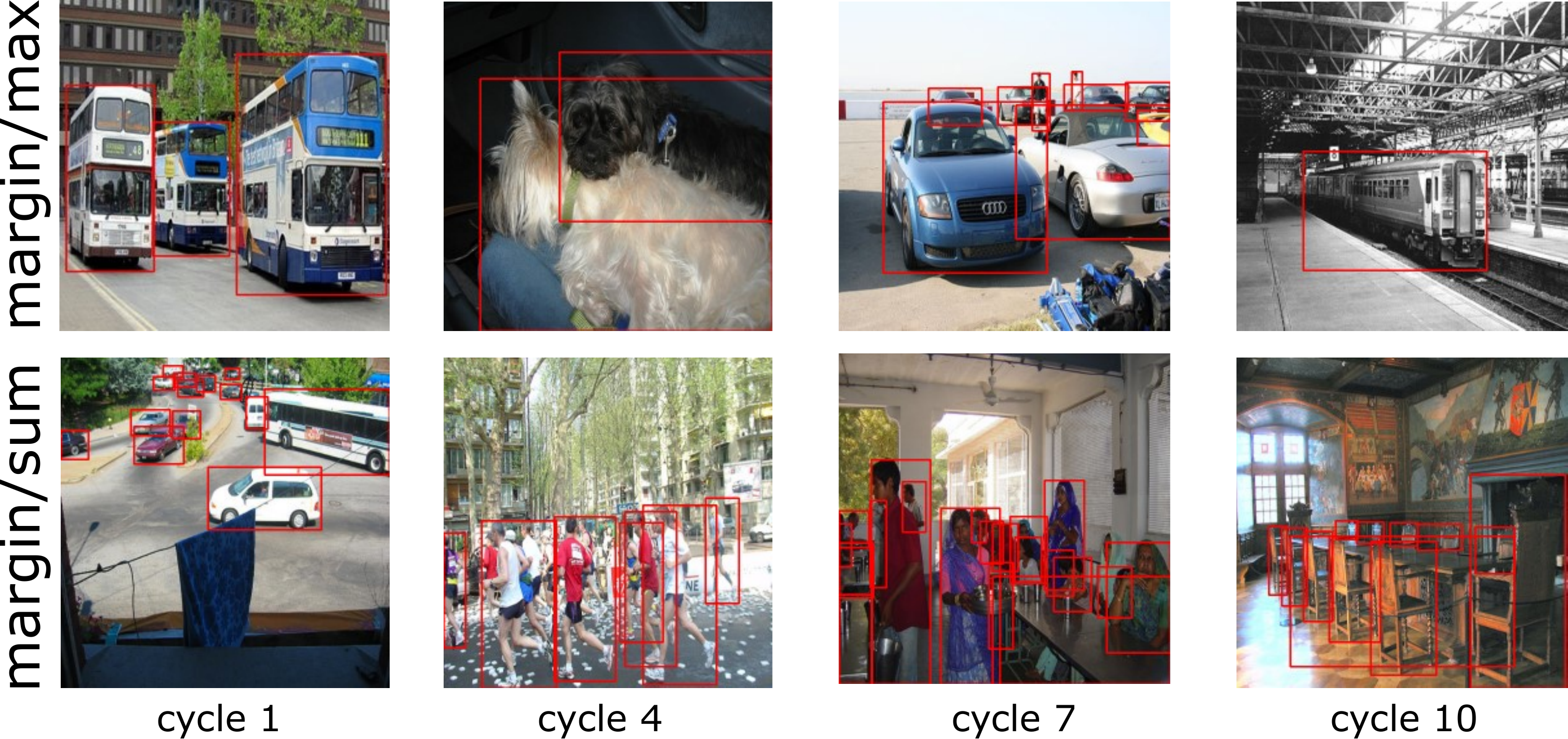}
    \caption{Qualitative results of models trained at different active learning cycles for two method combinations of our active learnign schemes. Bounding boxes are shown with red color irrespective of their category for simplicity.}
    \label{fig:qualresults}
\end{figure}
To understand the method's behaviour, we show some examples selected by our margin technique in \figurename~\ref{fig:qualresults}.
The sum accumulation method clearly results in images being selected that contain a large number of objects. Therefore, we observe empirical evidence for
our reasoning in \sectionname~\ref{sec:accumulation}.

\myparagraph{Comparison with non-zero-cost methods}
\begin{figure}[tbp]
    \centering
    \includegraphics[width=0.99\linewidth]{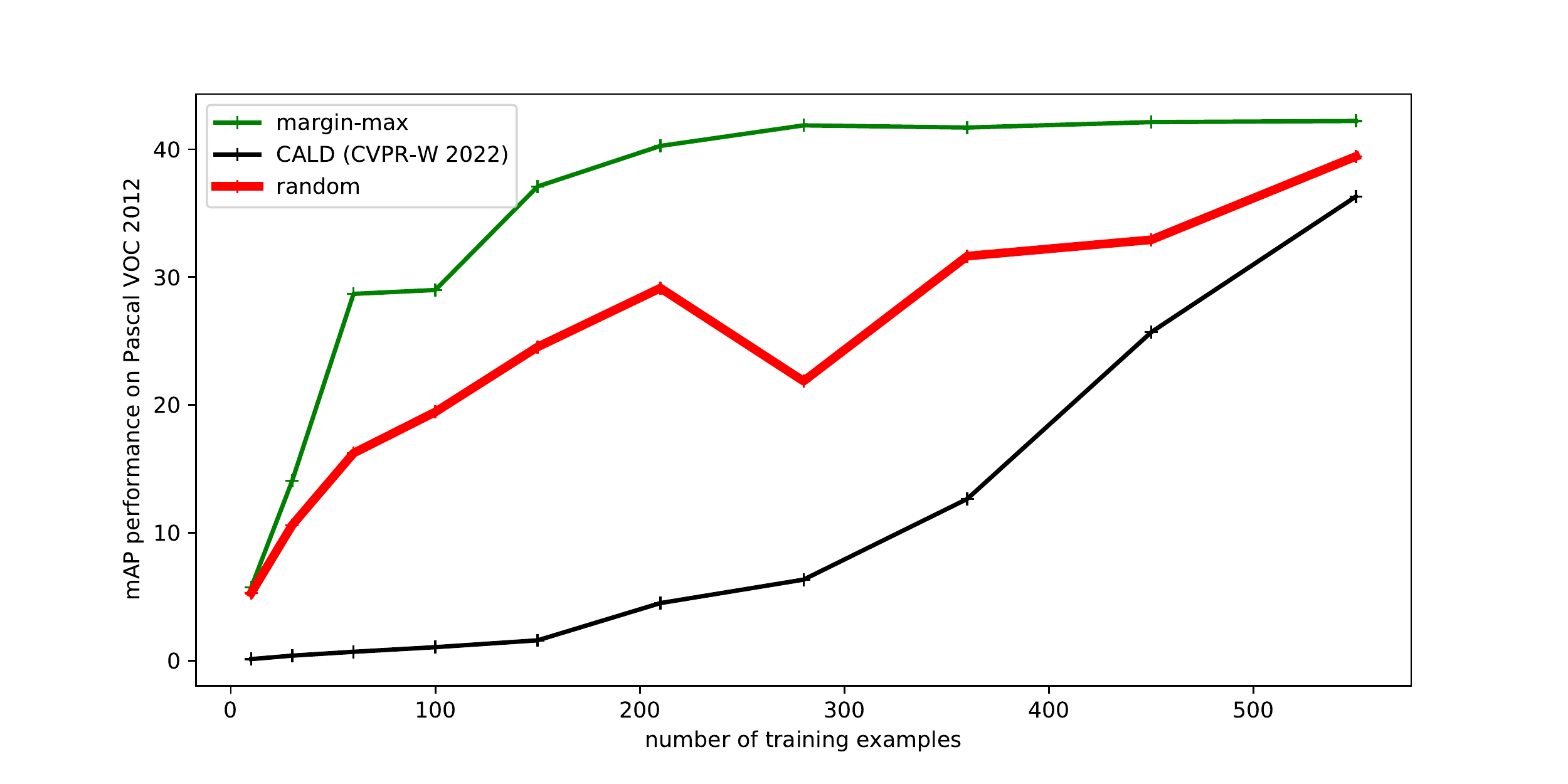}
    \caption{Comparison of our best zero-cost active learning method, the random baseline and the active learning method of \cite{yu2022consistency}.}
    \label{fig:soa}
\end{figure}
We also compare the best zero-cost active learning technique ``margin-max'' with CALD~\cite{yu2022consistency}, a recently published active learning method that measures
the relevance of an unlabeled image by augmenting the image and evaluating the stability of the detector results. A technique that involves a significant computational overhead.
The results are depicted in \figurename~\ref{fig:soa}. At first sight, the state-of-the-art method of \cite{yu2022consistency}
results in inferior performance even compared to standard passive learning (random). However, this is most likely due to the different
pretrained weights being used. In our case, we make use of models pretrained on the MS-COCO dataset~\cite{COCO}. In contrast, our
results of \cite{yu2022consistency} only use ImageNet classification weights. Furthermore, the results shown in the original
work of \cite{yu2022consistency} are based on models with a larger initial set of annotated examples. Further evaluation is necessary to allow for full comparision with
\cite{yu2022consistency}.

\section{Conclusion and Future Work}
In our paper, we evaluated active learning techniques that are easy to implement and do not involve a significant computational overhead besides model inference. In total, nine method combinations have been evaluated and compared.
Our best method achieved a significant performance gain compared to passive learning (no intelligent selection of unlabeled examples for annotation) and comes without additional cost.

Future work will concentrate on batch active learning with minimal computational overhead as well as integrating location uncertainty into the scoring functions. 

%
%
%
\newpage
\bibliographystyle{splncs04}
\bibliography{bibliography}
%
\end{document}